\documentclass{article}

\usepackage{microtype}
\usepackage{graphicx}
\usepackage{subcaption}
\usepackage{booktabs} 

\usepackage{hyperref}
\usepackage{algorithmic}
\usepackage{amsmath,amssymb,amsfonts}
\usepackage{multirow}
\usepackage{booktabs}
\usepackage{utfsym}
\usepackage{diagbox}
\usepackage[table]{xcolor}  
\definecolor{lightpurple}{HTML}{F3E8FF}

\usepackage[preprint]{icml2026}

\usepackage{amsmath}
\usepackage{amssymb}
\usepackage{mathtools}
\usepackage{amsthm}
\usepackage{caption}

\usepackage[capitalize,noabbrev]{cleveref}
\theoremstyle{plain}

\theoremstyle{definition}

\theoremstyle{remark}

\usepackage[textsize=tiny]{todonotes}

\icmltitlerunning{iPEAR: Iterative Pyramid Estimation with Attention and Residuals for Deformable Medical Image Registration}
\begin{document}

\twocolumn[
  \icmltitle{iPEAR: Iterative Pyramid Estimation with Attention and Residuals for Deformable Medical Image Registration}




  \begin{icmlauthorlist}
    \icmlauthor{Heming Wu}{csw}
    \icmlauthor{Di Wang}{ntu}
    \icmlauthor{Tai Ma}{zju}
    \icmlauthor{Peng Zhao}{ccst}
    \icmlauthor{Yubin Xiao}{ccst}
    \icmlauthor{Zhongke Wu}{bnu}
    \icmlauthor{Xing-Ce Wang}{bnu}
    \icmlauthor{Xuan Wu}{ccst}
    \icmlauthor{You Zhou}{csw,ccst}
  \end{icmlauthorlist}

  \icmlaffiliation{csw}{College of Software, Jilin University}
  \icmlaffiliation{ntu}{Joint NTU-UBC Research Centre of Excellence in Active Living for the Elderly, Nanyang Technological University}
  \icmlaffiliation{zju}{College of Computer Science and Technology, Zhejiang University}
  \icmlaffiliation{ccst}{Key Laboratory of Symbolic Computation and Knowledge Engineering of Ministry of Education, College of Computer Science and Technology, Jilin University}
  \icmlaffiliation{bnu}{School of Artificial Intelligence, Beijing Normal University}

  \icmlcorrespondingauthor{Xuan Wu}{wuuu22@mails.jlu.edu.cn}
  \icmlcorrespondingauthor{You Zhou}{zyou@.jlu.edu.cn}


  \icmlkeywords{Machine Learning, ICML}

  \vskip 0.3in
]



\printAffiliationsAndNotice{}  

\begin{abstract}
Existing pyramid registration networks may accumulate anatomical misalignments and lack an effective mechanism to dynamically determine the number of optimization iterations under varying deformation requirements across images, leading to degraded performance. To solve these limitations, we propose \textbf{iPEAR}. Specifically, iPEAR adopts our proposed Fused Attention-Residual Module (FARM) for decoding, which comprises an attention pathway and a residual pathway to alleviate the accumulation of anatomical misalignment. We further propose a dual-stage Threshold-Controlled Iterative (TCI) strategy that adaptively determines the number of optimization iterations for varying images by evaluating registration stability and convergence. Extensive experiments on three public brain MRI datasets and one public abdomen CT dataset show that iPEAR outperforms state-of-the-art (SOTA) registration networks in terms of accuracy, while achieving on-par inference speed and model parameter size. Generalization and ablation studies further validate the effectiveness of the proposed FARM and TCI.
\end{abstract}

\section{Introduction}

In single-modality settings, deformable image registration aims to align the corresponding anatomical structures between a fixed and moving image pair within the same imaging modality by estimating a dense non-linear deformation field, with widespread real-world applications in medical image analysis \cite{b1,b2}. To estimate the deformation field, conventional methods perform separate optimizations for each image pair, incurring high computational costs \cite{b9}.

To improve efficiency, deep learning methods have been widely adopted for deformable image registration, with pioneering works \cite{b10,b18} mainly adopting U-Net and Cascaded frameworks to estimate deformation fields and warp moving images. U-Net networks typically predict a single deformation field at the output layer, which limits their ability to model large and complex deformations \cite{b15}. Cascaded schemes address this issue by employing multiple subnetworks at different scales to progressively refine the deformation field, but they often re-extract features redundantly and underutilize previously encoded representations \cite{b26}.

\begin{figure}[!t]
    \centering
    \includegraphics[width=1.0\linewidth]{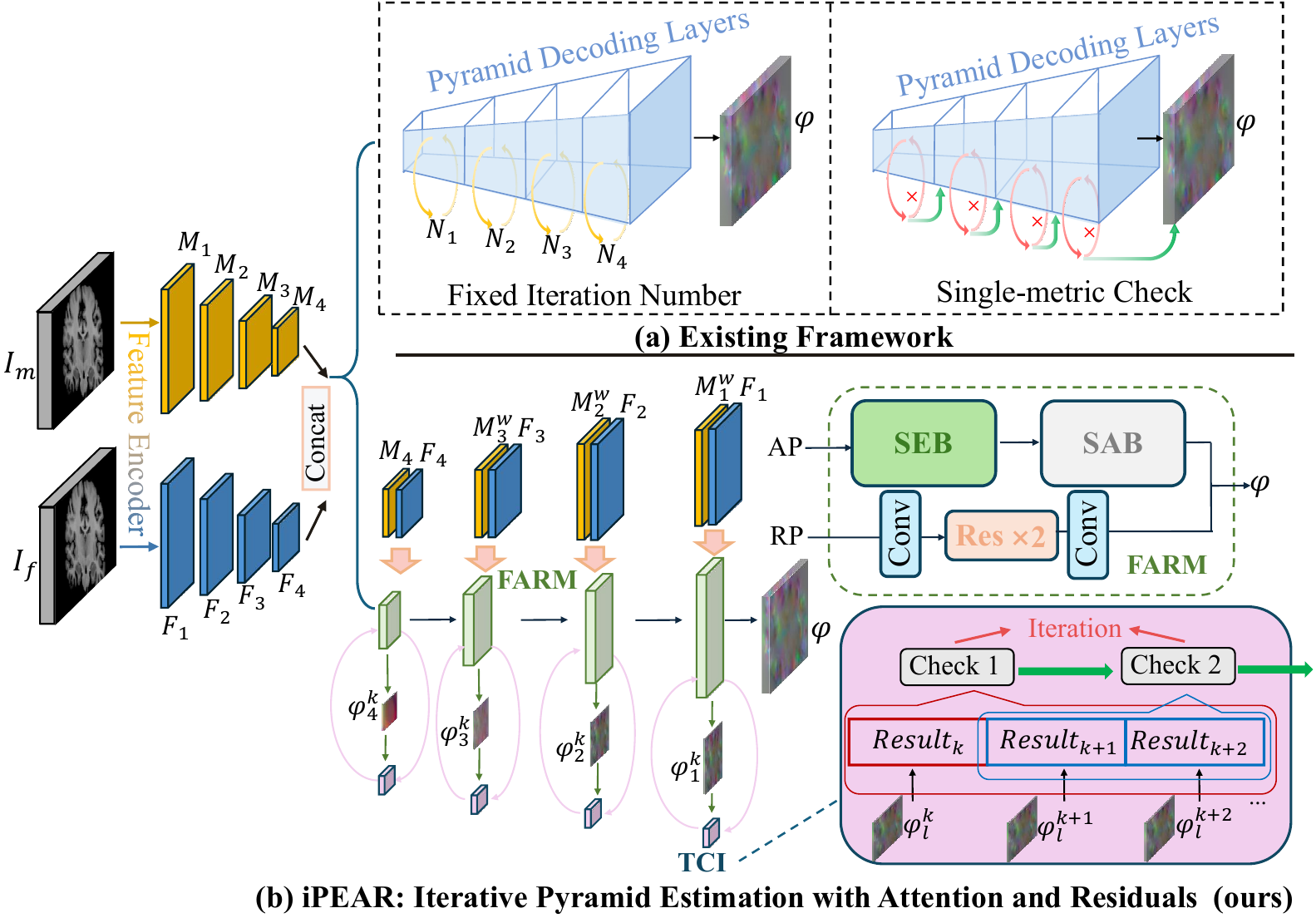}
    \caption{Illustration of the differences in decoding processes between existing pyramid models and our proposed iPEAR. Sub-figure~(a) portrays the existing models \cite{b26} (left) and \cite{b24} (right), which do not specifically prevent propagation and accumulation of anatomical structure misalignments nor adaptively determine the number of iterations. Sub-figure~(b) depicts our proposed iPEAR. At each decoding layer, iPEAR employs our designed FARM to learn to suppress irrelevant features, thereby reducing misalignment accumulation. Subsequently, iPEAR employs our novel dual-stage TCI strategy to adaptively determine the number of iterations for guiding FARM in the progressive optimization process of the deformation field.}
    \label{fig:1}
    \vspace{-0.2cm}
\end{figure}

To minimize unnecessary feature extraction, recent studies have proposed pyramid models \cite{b31,b23} that employ a single encoder to extract multi-scale feature maps. Subsequently, at each scale, the decoding layer fuses the feature maps extracted from both the fixed and moving image branches to estimate the deformation field. The estimated deformation field is then passed to the subsequent decoding layer (see Figure~\ref{fig:1}(a)), thereby achieving coarse‑to‑fine estimation. Although pyramid architectures facilitate multi-scale feature extraction, their decoders may exacerbate the propagation and accumulation of anatomical structure misalignments, especially those stemmed from irrelevant information in deep, coarse‑scale features. This exacerbation ultimately constrains the overall registration performance \cite{b30}. In addition, existing single-modality pyramid models often adopt vanilla residual decoders, which limit their performance (see Tables~\ref{tab:1} and \ref{tab:flare}). Moreover, in scenarios involving large deformations, a one-pass estimation per scale typically fails to yield an accurate deformation field, necessitating iterative estimations at each scale. Consequently, SOTA models perform multiple estimations at each scale to iteratively optimize the deformation fields. For example, Wang et al. \yrcite{b26} adopted a fixed number of iterations, while Ma et al. \yrcite{b24} introduced a single-stage mechanism to dynamically determine the number of iterations at each scale. However, due to the diversity among image pairs, neither a fixed number of iterations nor a single-stage criterion may effectively adapt to varying deformation requirements, often resulting in premature termination or excessive iterations that degrade registration performance (see Section~\ref{sec4.1}). 

To mitigate misalignment propagation and accumulation in pyramid architectures, we propose the Fused Attention-Residual Module \textbf{(FARM)} as the core component of the decoding layer to emphasize informative features. Specifically, FARM follows a dual-pathway design, including an Attention Pathway (AP) and a Residual Pathway
(RP) (see Figure~\ref{fig:1}~(b)). The AP comprises a Squeeze Excitation Block (SEB) and a Spatial Attention Block (SAB). SEB applies channel attention to suppress irrelevant features and alleviate accumulated anatomical misalignment, while SAB further highlights salient spatial regions based on the refined feature maps, thereby reducing misalignment accumulation. In parallel, RP employs two convolutional layers for channel transformation and two sequential residual blocks to extract fine-grained anatomical details, supporting comprehensive and precise structural modeling (see Section~\ref{sec3.2}). By adopting this design, FARM achieves high-level registration performance compared to a vanilla residual decoder without substantially increasing the number of learnable parameters. Moreover, to adaptively determine the number of optimization iterations for diverse image pairs, we propose the dual-stage Threshold-Controlled Iterative \textbf{(TCI)} strategy, which employs a sliding window of size $t$ to dynamically track the estimated deformation fields and quantifies each by the similarity difference between the warped moving image $I_m^w$ and the fixed image $I_f$. TCI then assesses the stability of deformation (i.e., first stage) by computing the standard deviation $\varepsilon_l$ of all similarity differences within this window. Upon reaching stability (see Section~\ref{subsec:C}), TCI further evaluates convergence (i.e., second stage) by measuring the change $\Delta s$ between similarity differences of the two most recently estimated deformation fields within this window. TCI is designed to be efficient to enable adaptive determination of the number of iterations for each scale of the underlying image pair, without significantly increasing registration time (see Section~\ref{sec4.1}). We coin the network that integrates FARM and TCI as the iterative Pyramid Estimation with Attention and Residuals \textbf{(iPEAR)} network. 
\textbf{To the best of our knowledge, iPEAR is the first pyramid network to jointly model channel and spatial attention alongside residual blocks in its decoder, while dynamically determining the number of optimization iterations.}

The key contributions of this work are as follows:

\textbf{I)} We propose a dual-pathway FARM for decoding, comprising AP and RP, where AP suppresses irrelevant information and RP extracts fine-grained anatomical details. 

\textbf{II)} We propose the dual-stage TCI strategy, which monitors the stability and convergence of recent registration results to adaptively determine the number of optimization iterations for image pairs with varying deformation requirements.

\textbf{III)} To assess the performance of iPEAR, we conduct experiments on three public brain Magnetic Resonance Imaging (MRI) datasets spanning different deformation scales and one public abdomen Computed Tomography (CT) dataset. The experimental results demonstrate that iPEAR outperforms SOTA registration models in accuracy across all datasets, while achieving on-par inference speed and model parameter size. Moreover, generalization and ablation studies further validate the effectiveness of FARM and TCI.

\section{Related Work}

\begin{figure*}[!t]
    \centering
    \vspace{-0.2cm}
\includegraphics[width=0.8\textwidth]{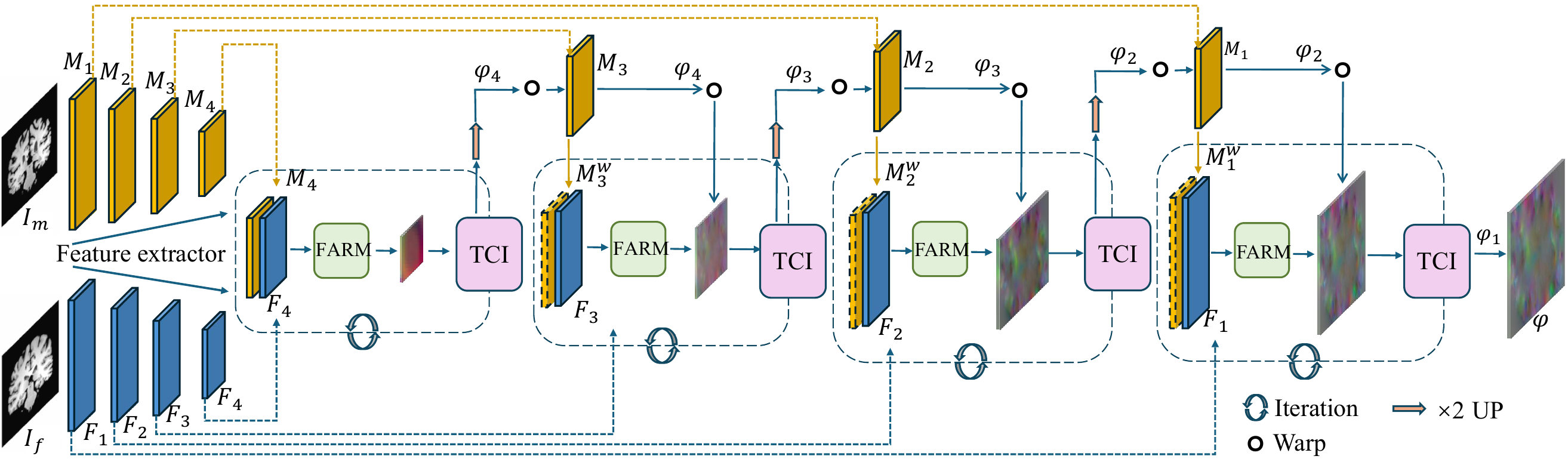}
    \caption{Overall iPEAR model architecture. A weight-sharing encoder first extracts multi-scale feature maps \(\{F_l\}\) and \(\{M_l\}\) for the fixed image $I_f$ and the moving image $I_m$, respectively. In the subsequent decoding layers, iPEAR employs FARM to emphasize informative features and estimate the deformation field $\varphi_l$, while TCI adaptively determines the number of iterations for each decoding layer, enabling progressive optimization of $\varphi_l$. Notably, the encoder processes higher-resolution feature maps first ($F_1 \rightarrow F_4, M_1 \rightarrow M_4$), while the decoder begins deformation estimation on the lowest‑resolution maps and proceeds to finer scales ($F_4 \rightarrow F_1, M_4 \rightarrow M_1$).}
    \label{fig:2}
\vspace{-0.5cm}
\end{figure*}
In this section, we review the relevant literature.

\textbf{Deep Learning (DL) Registration Models:} DL models have become the mainstream for solving registration tasks due to their advantages in inference speed and accuracy compared to conventional methods \cite{b11}. Pioneer DL models primarily employed U-Net architectures, which estimate only a single deformation field at the final output layer and consequently struggle to accommodate complex deformations \cite{b48}. To overcome this limitation, certain studies \cite{b18,b20,b21} proposed cascaded registration networks, which employ multiple sub-networks to extract features at distinct resolutions and estimate the corresponding deformation fields. Although these models achieve moderate accuracy improvements, they perform repeated encoding and decoding, leading to redundant feature extraction and elevated computing overhead \cite{b24}.
To minimize redundant feature processing, recent studies \cite{b23} proposed the pyramid network, which adopts a single encoder followed by multiple decoding layers to estimate hierarchical deformation fields for continuous coarse‑to‑fine warping images. Although this architecture facilitates effective multi-scale feature extraction and progressive deformation estimation, it may exacerbate the propagation and accumulation of anatomical structure misalignments, especially those stemmed from irrelevant information in deep, coarse‑scale features, ultimately affecting the overall registration performance \cite{b30}. To better overcome this limitation, we propose FARM to suppress irrelevant features in the decoding layers, thereby mitigating misalignment propagation. Moreover, FARM excels at capturing anatomical details, ensuring the accuracy and consistency of deformation fields at each scale (see Section~\ref{sec3.2}).
\noindent\textbf{Iterative Strategy in Pyramid Models:} For small deformations, prior pyramid models \cite{b31} estimated only a single deformation field per decoding layer (each corresponding to one scale). However, under large deformations, a single estimation per scale is often insufficient to produce an accurate deformation field. Consequently, each scale requires iterative estimations to refine the deformation field. For example, Wang et al. \yrcite{b26} proposed RDP, which performs a predefined number of iterations for each scale to refine the deformation field. Inevitably, this fixed number of iterations lacks adaptability to varying deformation requirements. To adaptively determine the number of iterations, Ma et al. \yrcite{b24} proposed a single-stage mechanism, which evaluates convergence by monitoring differences between two successive registration results. However, relying on a single-metric check only captures partial information and may lead to premature iteration termination, potentially compromising the overall registration accuracy (see Tables~\ref{tab:6} and \ref{tab:iter_of_TCI}). To dynamically determine when to stop iterative optimization, we propose a dual-stage mechanism, which evaluates the stability of historical registration results and convergence subsequently  (see Section~\ref{subsec:C} for more details).



\section{iPEAR Dynamics }

The architecture of \textbf{iPEAR} is schematically illustrated in Figure~\ref{fig:2}. In this section, we first introduce \textbf{FARM}, which extracts anatomical structure details and learns to suppress irrelevant features to mitigate misalignment. We then introduce the dual-stage \textbf{TCI} strategy, which adaptively determines the number of iterations for optimization. 

\textbf{Overview of iPEAR:} iPEAR comprises a multi-scale encoder and a multi-layer decoder. 
iPEAR first employs a weight-sharing feature encoder to extract multi-scale feature map sets \(\{F_l\}\) and \(\{M_l\}\), where $l\in\{1,...,4\}$ (following prior arts in this field: \cite{b31,b24}), from the fixed image $I_f$ and moving image $I_m$, respectively. Specifically, iPEAR adopts the encoder used in \cite{b48}, which comprises four hierarchical convolutional encoder blocks. Each block includes a 3D convolutional layer followed by Neighborhood Attention \cite{b51} (except for the first block), and then applies average pooling to halve the spatial dimensions of the feature maps.
Then, for the $l$th scale (i.e., $l$th decoding layer), iPEAR applies FARM to process \(\{F_l\}\) and \(\{M_l\}\), extracting anatomical structure details and learning to suppress irrelevant features, to estimate the corresponding deformation fields (see Section~\ref{sec3.2}). To adaptively determine the number of iterations in each decoding layer, iPEAR relies on TCI, which monitors the similarity difference of the recently estimated registration results (see Section~\ref{subsec:C}). 


\begin{figure}[!t]
    \centering
    \vspace{-0.2cm}\includegraphics[width=1\linewidth]{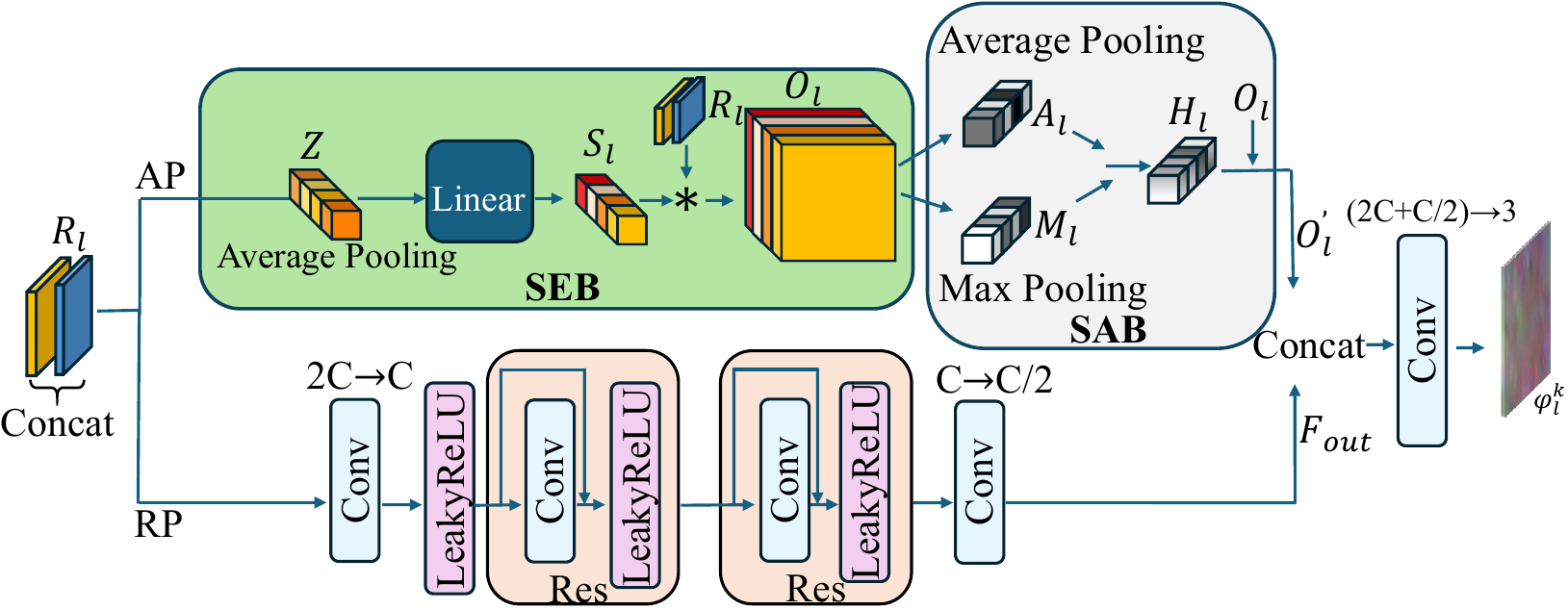}
    \caption{The proposed FARM adopts a dual-pathway architecture. The AP is composed of SEB and SAB. SEB applies channel attention to suppress irrelevant features and alleviate accumulated anatomical misalignment. Building on these refined feature maps, SAB emphasizes salient spatial regions, which further reduces misalignment accumulation. In parallel, RP includes two convolutional layers for channel transformation and two sequential residual blocks that extract fine-grained anatomical details, supporting comprehensive and precise structural modeling.}
    \label{fig:3}
    \vspace{-0.2cm}
\end{figure}

\subsection{Fused Attention-Residual Module (FARM)}
\label{sec3.2}

Conventional residual convolutional structures, while effective at preserving anatomical consistency and modeling local details, typically apply uniform weighting across feature channels. This limits the network’s ability to emphasize subtle yet critical deformation cues that are essential for high-accuracy registration \cite{b24}. In contrast, attention mechanisms excel at adaptively highlighting discriminative features and suppressing redundant information. However, when used alone, they often lack sufficient capability for robust local structure modeling and stable deformation estimation \cite {b14}. Therefore, an effective decoder should combine the local modeling strength of residual convolutions with the adaptive feature selection of attention mechanisms to achieve high-level registration performance. 

To this end, we propose FARM as the core component of the decoder for estimating deformation fields, which follows a dual-pathway design consisting of an Attention Pathway (\textbf{AP}) and a Residual Pathway (\textbf{RP}), as shown in Figure~\ref{fig:3}. Specifically, when $l=4$ and $k=1$, where $k$ denotes the current iteration number at the $l$th layer, FARM directly takes $F_4$ and $M_4$ as inputs to estimate the deformation field $\varphi_4^1\in\mathbb{R}^{C_\varphi\times H \times W \times D}$ via its dual-pathway architecture, where $C_\varphi$, $H$, $W$, and $ D$ denote the number of channels, height,
width, and depth of the deformation field, respectively. When $l \in\{1,2,3\}$ and $k=1$, FARM adopts the deformation field $\varphi_{l+1}$ of the $(l+1)$th layer to warp $M_l$ to obtain $M_l^w$ ($M_l^w= M_l \circ \varphi_{l+1}$), where $\circ$ denotes the deformation operation, implemented via the Spatial Transformer Network \cite{b41}. Then, FARM utilizes $F_l$ and $M_l^w$ to estimate the deformation field $\varphi_l^1\in\mathbb{R}^{C_\varphi\times H \times W \times D}$ for the $l$th layer through its dual pathways. Otherwise, FARM employs ${\varphi}_{l}^{k-1}$ of the $(k-1)$th iteration to warp $M_l$ to obtain $M_l^w$ ($M_l^w= M_l \circ \varphi_l^{k-1}$). Formally,
\begin{equation}\varphi_{l}^k=
\text{FARM}(R_l),
\end{equation}
\begin{equation}R_{l}=\small{\left\{\begin{array}{ll}
\left(|M_{4}, F_{4}|\right), & l=4 \land k=1, \\
\left(|M_l^w, F_{l}|\right) , & \text{Otherwise}, \\
\end{array}\right.}
\end{equation}
where $|\cdot|$ denotes the concatenation operator, and the final expression of $\varphi_l^k$ is given later in (\ref{phi}).

AP consists of a Squeeze-and-Excitation Block (SEB) and a Spatial Attention Block (SAB), while RP comprises two convolutional layers for channel transformation and two sequential residual blocks. Specifically, AP extends the 2D SEB \cite{SEB} to a 3D variant to suppress irrelevant features in $R_l$, thereby mitigating propagation and accumulation of anatomical misalignment. The 3D SEB first employs average pooling to compress spatial dimensions and extract global informative features, producing the channel descriptor vector $Z_l \in\mathbb{R}^{2C}$ as follows:
\begin{equation}Z_{l,c}=\frac{1}{H\cdot W\cdot D}\sum\nolimits_{i=1}^H\sum\nolimits_{j=1}^W\sum\nolimits_{d=1}^DR_{l,c,i,j,d},
\end{equation} 
where $C$ denotes the number of channels of the image and $c \in \{1, \cdots, 2C\}$ . Then, SEB applies two fully connected layers with LeakyReLU $\gamma$ and sigmoid $\sigma$ activation functions to compute the channel weight vector $S_l$ as follows:
\begin{equation}
S_l=\sigma({W}_2 \gamma({W}_1 Z_l)),
\end{equation}
where ${W}_1\in\mathbb{R}^{\frac{2C}{r}\times 2C}$ and ${W}_2\in\mathbb{R}^{ 2C\times\frac{2C}{r}}$. The hyperparameter $r$ denotes the reduction ratio, which is used to gauge the computational cost. The first linear layer reduces the dimensionality of $Z_l$ to $\frac{2C}{r}$, for efficient channel-wise compression and information aggregation. The second linear layer then restores the dimensionality back to $2C$. By adopting this design, SEB can model inter-channel dependencies and recalibrate channel responses. Consequently, SEB adaptively emphasizes deformation-sensitive channels while suppressing less informative ones, mitigating the tendency of uniform channel weighting to overlook subtle yet critical deformation cues  (see Table~\ref{tab:2}). Subsequently, the resulting weight vector $S_l$ is applied to $R_l$ via per‑channel element‑wise multiplication~$\ast$, yielding the enhanced feature map $O_{l} \in\mathbb{R}^{2C\times H \times W \times D}$ as follows:
\begin{equation}O_{l,c}=R_{l,c}\ast S_{l,c}, c \in \{1, \cdots, 2C\}.\end{equation}

To further alleviate misalignment accumulation, SAB applies average pooling and max pooling to the enhanced feature map $O_l$ to obtain descriptor vectors $A_{l}$ and $M_{l}$, respectively. These two descriptors are then fused through a convolutional layer to produce the hybrid vector $H_l$, which captures complementary spatial cues. Formally,
\begin{equation}
H_l =Conv(|A_{l}, M_{l}|) ,\end{equation}
\begin{equation}A_{l,c}=\frac{1}{H\cdot W\cdot D}\sum\nolimits_{i=1}^H\sum\nolimits_{j=1}^W\sum\nolimits_{d=1}^DO_{l,c,i,j,d},
\end{equation} 
\begin{equation}M_{l,c}=\max_{i=\{1,\ldots,H\},j=\{1,\ldots,W\},d=\{1,\ldots,D\}}O_{l,c,i,j,d}.\end{equation}
Subsequently,
the descriptor vector $H_l$ is applied to $O_l$ via per‑channel element‑wise multiplication $\ast$, yielding the hybrid feature map $O^{'}_{l} \in\mathbb{R}^{2C\times H \times W \times D}$ as follows:
\begin{equation}O{'}_{l,c}=O_{l,c}\ast H_{l,c}, c \in \{1, \cdots, 2C\}.\end{equation}

RP comprises a 3D convolutional layer ${Cl}_1$, two residual blocks ($Res_1$ and $Res_2$), and a final 3D convolutional layer $Cl_2$. Specifically, $R_l$ is first fed into a 3D convolutional layer, which adjusts the channel dimension from $2C$ to $C$, followed by a LeakyReLU function $\gamma$. The resulting features then pass through $Res_1$ and $Res_2$, each consisting of a 3D convolutional layer followed by $\gamma$. Finally, $Cl_2$ outputs the feature map $F_{\text{out}}$ with $\frac{C}{2}$ channels.
Formally,
\begin{equation}Res_i(x)=\gamma (Conv(x)+x), i \in \{1, 2\},\end{equation}
\begin{equation}
F_{o}=Cl_2^{C\rightarrow \frac{C}{2}}(Res_2(Res_1(\gamma(Cl_1^{2C\rightarrow C}(R_l))))),
\end{equation}
where $x$ denotes the input to the residual block.
After obtaining $O^{'}_{l}$ and $F_{o}$, FARM finally applies a convolutional layer $Cl_3$ to combine them and estimate the intermediate deformation field $\overline{\varphi}_l^k$ for the $k$th iteration as follows: 
\begin{equation}\overline{\varphi}_l^k={Cl_3^{(\frac{C}{2}+2C\rightarrow 3)}}(|O^{'}_{l},F_{o}|).
\label{eq:phi}\end{equation}

After obtaining $\overline{\varphi}_l^k$,  FARM forms the final $\varphi_l^k$ for the $k$th iteration by composing it with the deformation field $\varphi_l^{k-1}$ of the preceding iteration when $k\neq 1$. For the first iteration (i.e., $k=1$) with  $l \neq 4$, FARM uses the final deformation field from the $(l+1)$th layer $\varphi_{l+1}$, in place of $\varphi_l^{k-1}$. When $l=4$ and $k=1$, FARM simply sets $\varphi_l^{k}=\overline{\varphi}_l^{k}$.   Formally,
\begin{equation}\varphi_l^k=
\left\{\begin{array}{ll}
\varphi_l^{k-1}\circ\overline{\varphi}_l^k+\overline{\varphi}_l^k, & k\neq1, \\
\varphi_{l+1}\circ\overline{\varphi}_l^k+\overline{\varphi}_l^k, & k = 1, l \neq 4, \\
\overline{\varphi}_l^k, & \text{Otherwise.}
\end{array}\right.
\label{phi}
\end{equation}
Finally, when TCI terminates at the $k$th iteration, FARM upsamples $\varphi_l^k$ with a factor of 2 to obtain the final deformation $\varphi_l$ for the $l$th layer.   
Notably, unlike the decoder adopted by IIRP \cite{b24}, which relies solely on multiple sequential residual blocks to estimate the deformation field, FARM integrates channel and spatial attention in parallel with a small number of residual blocks, designed to achieve enhanced performance through richer features (see Table~\ref{tab:1}). In addition, the results further validate that AP and RP effectively suppress irrelevant features and capture fine-grained anatomical details, respectively (see Appendix~\ref{appendix_grad} for visualizations of feature attention patterns in AP and RP).



\subsection{Threshold-Controlled Iterative (TCI) Strategy }\label{subsec:C}
\begin{figure}[h]
    \centering
    \includegraphics[width=0.46\textwidth]{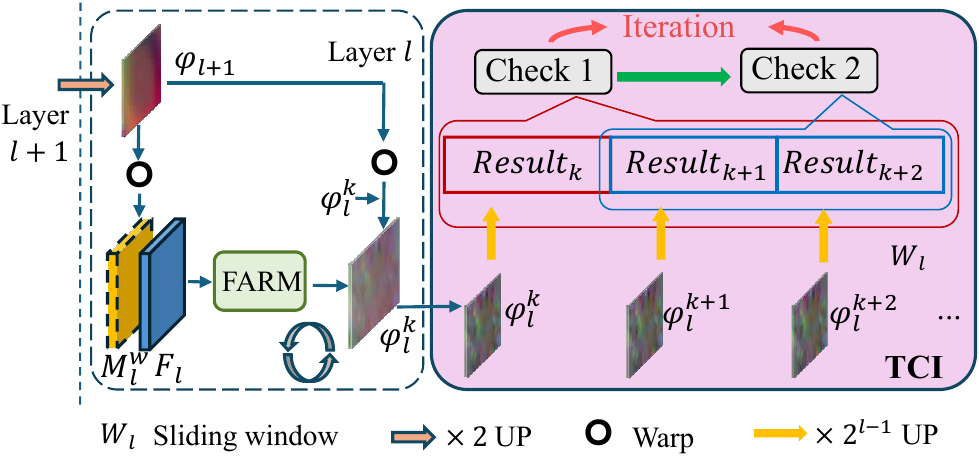}
    \caption{Depiction of the proposed TCI strategy, which determines when to stop the iteration through a dual-stage mechanism. TCI first checks whether the std $\varepsilon_l$ of similarity differences across all historical registration results in $W_l$ is lower than a threshold $\delta_s$. If yes, TCI further computes the change $\Delta s$ between similarity differences of the two most recent registration results in $W_l$. If $\Delta s$ is smaller than another threshold $\delta_c$, it is considered that the appropriate stopping criteria have been reached.}
    \label{fig:4}
    \vspace{-0.4cm}
\end{figure}


The number of iterations is crucial for optimizing the deformation field; early termination may yield suboptimal accuracy, while excessive iterations may introduce misalignment and consume unnecessary computing resources. 
To adaptively determine the number of iterations for varying image pairs, we propose a dual-stage TCI strategy to evaluate both stability and convergence using historical registration results to dynamically determine when to stop. As shown in Figure~\ref{fig:4}, at the $l$th decoding layer, TCI employs a sliding window $W_l$ to dynamically track the similarity differences of $t$ number of recently estimated registration results.

To compute the similarity difference $s_l^k$ between the warped image $I_m^w$ and the fixed image $I_f$ at the $k$th iteration, TCI first upsamples the deformation field $\varphi_l^k$, yielding the upsampled deformation field $\tilde\varphi_l^k$ at the original moving image resolution.
The upsampling is implemented via linear interpolation with a factor of $2^{l-1}$. Next, $\tilde{\varphi}_l^k$ provides a displacement vector for each voxel to be applied to the moving image $I_m$ to produce the corresponding warped image $I_m^w$, i.e., $I_m^w = I_m\circ\tilde{\varphi}_l^k$. Formally, 
\begin{equation}
s_l^k=\text{sim}\left(I_f,I_m^w\right),
\label{eq:sim}
\end{equation}
where $\mathrm{sim}(\cdot)$ denotes the similarity metric. In this work, we always adopt Normalized Cross-Correlation (NCC) to compute the registration similarity after examining various candidates (see Appendix~\ref{appendix3}). 

After obtaining the similarity differences of $t$ number of recently estimated registration results, TCI evaluates their stability and convergence in a dual-stage process, respectively. In the first stage, TCI assesses stability by computing the standard deviation $\varepsilon_l$ of similarity differences $s_l^k$ within the window $W_{l}=\{s_l^{k-t+1},\cdots,s_l^k\}$ as follows:
\begin{equation}
\label{eq:std}
\varepsilon_l=\text{std}\left(W_{l}\right).
\end{equation}
If $\varepsilon_l$ falls below the predefined stability threshold $\delta_s$, the most recent $t$ number of deformation fields are deemed stable. TCI then proceeds to the subsequent convergence evaluation stage. Otherwise, another iteration is performed without the necessity to continue the subsequent stage.

\begin{algorithm}[!t]
\caption{Registration process in the $l$th decoding layer}
\label{alg}
\begin{algorithmic}[1]
\STATE \textbf{Input:} $(l+1)$th deformation field ${\varphi}_{l+1}$, feature maps $F_l$, $M_l$, original images $I_f$, $I_m$, $W_l$, $k \leftarrow 1$
\STATE \textbf{Output:} $l$th deformation field ${\varphi}_l$

\STATE $\varphi_l^{k-1} \leftarrow {\varphi}_{l+1}$ \textcolor{gray}{\hfill \#If $k \neq 1 \lor l=4$, this step is skipped} 
\WHILE{$k\leq k_{max}$}
    \STATE $M_l^w \leftarrow M_l \circ \varphi_l^{k-1}$\textcolor{gray}{\small{\#If $k = 1\land l=4$, this step is skipped}}
    \STATE $R_l \leftarrow \left(|M_{l}^{w}, F_{l}|\right)$ \textcolor{gray}{  \small{\#If $k=1\land l=4 $, $R_4 \leftarrow \left(|M_{4}, F_{4}|\right)$}}
    \STATE $O_l \leftarrow \text{\textbf{SEB}}(R_l)$, \quad $H_l = \text{\textbf{SAB}}(|O_l|)$  
    \STATE $O^{'}_l \leftarrow O_l \ast H_l$ \textcolor{teal}{\hfill\#AP output} 
    \STATE $F_{o} \leftarrow {Cl}_2(Res_2(Res_1(\gamma({Cl}_1(R_l)))))$ \textcolor{teal}{\hfill\#RP output} 
    \STATE $\overline{\varphi}_l^k = {Cl}_3(|O^{'}_l, F_{o}|)$ \textcolor{gray}{\hfill\#deformation estimation}
    \STATE $\varphi_l^k \leftarrow {\varphi}_l^{k-1} \circ \overline{\varphi}_l^k + \overline{\varphi}_l^k$ \textcolor{gray}{\#If $l = 4 \land k=1$, this step is skipped.}
    \STATE $\tilde\varphi_l^k \leftarrow \text{upsample}(\varphi_l^k,2^{l-1})$
    \STATE $s_l^k \leftarrow \text{sim}(I_f, I_m \circ \tilde\varphi_l^k)$ \textcolor{gray}{\hfill\#see (\ref{eq:sim})}
    \STATE append $s_l^k$ to $W_{l}$ \textcolor{gray}{\#If $\text{len}(W_l)=t+1$, remove $s_l^{k-t}$ from $W_{l}$}
    \STATE \textbf{if} $\text{len}(W_l) \geq t$ \textbf{then}\textcolor{teal}{\hfill \#TCI invoked}
    \STATE \hspace{1em}$\varepsilon_l\leftarrow\text{std}\left(W_{l}\right)$ \textcolor{gray}{\hfill \#see (\ref{eq:std})}
    \STATE \hspace{1em}\textbf{if} $\varepsilon_l \leq \delta_s$ \textbf{then} \textcolor{gray}{ \hfill \#Stage 1: assess stability}
    \STATE \hspace{2em}$\Delta s\leftarrow s_l^k - s_l^{k-1}$ \textcolor{gray}{\hfill \#see (\ref{eq:delta})}
    \STATE \hspace{2em}\textbf{if} $\Delta s \leq \delta_c$ \textbf{then}  \textcolor{gray}{\hfill\#Stage 2: assess convergence}
    \STATE \hspace{3em}\textbf{return} ${\varphi}_{l} \leftarrow \text{upsample}(\varphi_l^k,2)$
    \STATE $k \leftarrow k + 1$
\ENDWHILE
\STATE \textbf{return} ${\varphi}_{l} \leftarrow \text{upsample}(\varphi_l^{k_{max}},2)$
\end{algorithmic}
\end{algorithm}
In the subsequent convergence evaluation stage, TCI further computes the change in similarity differences, denoted as  $\Delta s$, between the two most recent iterations within $W_l$, which is defined as follows: 
\begin{equation}
\Delta s = s_l^k - s_l^{k-1}.
\label{eq:delta}
\end{equation}
If $\Delta s$ falls below the predefined convergence threshold $\delta_c$, the deformation field at the $k$th iteration is deemed converged, the iteration terminates and the deformation field $\varphi_l$ is passed to the $(l-1)$th layer as input (see Figure~\ref{fig:2}). Otherwise, additional iterations continue at the $l$th layer. Following \cite{b24}, TCI also leverages a predefined maximum number of iterations $k_{max}$ to avoid rare non-convergence situations. The pseudocode of the overall registration process is presented in Algorithm~\ref{alg}.


Notably, our TCI fundamentally differs from IIRP’s single‑stage method \cite{b24}, which performs only one similarity check between two consecutive registrations and may prematurely terminate due to stochastic convergence. TCI adopts a two-stage evaluation with an additional stability check, enabling more informed stopping decisions while maintaining comparable inference time (see Table~\ref{tab:6}).

\begin{table*}[!t]
\centering
\caption{Performance comparison of different registration models on the Mindboggle, LPBA, and IXI datasets}
\renewcommand{\arraystretch}{1}
\setlength{\tabcolsep}{3.5pt}
\small
\vspace{-0.2cm}
\resizebox{0.95\linewidth}{!}{
    \begin{tabular}{c c ccc ccc ccc c}
\cmidrule[\heavyrulewidth]{2-12}
& \multirow{2}{*}{\textbf{Model}} & \multicolumn{3}{c}{\textbf{Mindboggle}} & \multicolumn{3}{c}{\textbf{LPBA}} & \multicolumn{3}{c}{\textbf{IXI}} & \multirow{2}{*}{\textbf{\#Param}↓} \\
\cmidrule(lr){3-5}
\cmidrule(lr){6-8}
\cmidrule(lr){9-11}
& & Dice(\%)↑ & Time(s)↓ & $|{J}_{\varphi}|_{\leq0}$↓ 
  & Dice(\%)↑ & Time(s)↓ & $|{J}_{\varphi}|_{\leq0}$↓ 
  & Dice(\%)↑ & Time(s)↓ & $|{J}_{\varphi}|_{\leq0}$↓ & \\ \cmidrule(lr){2-12}
\multirow{2}{*}{\rotatebox[origin=c]{90}{\scalebox{0.9}{U-Net}}} 
& VM (CVPR'18)    & 56.0 & \textbf{0.09} & $<$0.7\%
             & 65.1 & \textbf{0.02} & $<$0.3\%
             & 75.9 & \textbf{0.01} & $<$0.2\% & \textbf{319.4K} \\
& TM (MIA'22)    & 61.5 & 0.10 & $<$1.0\%
             & 67.0 & 0.10 & $<$0.78\%
             & 76.9 & 0.03 & $<$0.43\% & 45650.3K \\  \cmidrule(lr){2-12}
\multirow{2}{*}{\rotatebox[origin=c]{90}{\scalebox{0.9}{Cascaded}}} 
& RCN (ICCV'19)         & 63.4 & 0.19 & $<$0.8\%
             & 71.4 & 0.05 & $<$0.007\%
             & 78.6 & 0.06 & $<$0.01\% & 958.1K \\         
& SDH (TMI'23)      & 63.6 & 0.82 & $<$0.2\%
             & 70.5 & 0.77 & $<$0.4\%
             & 79.3 & 0.08 & $<$0.2\% & 17862.0K \\ \cmidrule(lr){2-12}
\multirow{8}{*}{\rotatebox[origin=c]{90}{\scalebox{0.9}{Pyramid}}} 
& RDN (JBHI'22)           & 62.5 & 0.08 & $<$0.03\%
             & 70.6 & 0.04 & $<$0.05\%
             & 78.8 & 0.02 & $<$0.06\% & 28653.2K \\
& PR++ (MIA'22)  & 63.1 & 0.47 & $<$0.5\%
             & 70.4 & 0.64 & $<$0.07\%
             & 79.3 & 0.07 & $<$0.06\% & 1208.0K \\
& I2G (MMMI'22)   & 62.5 & 0.51 & $<$0.03\%
             & 71.7 & 0.06 & $<$0.006\%
             & 79.3 & 0.51 & $<$0.02\% & 865.2K \\
& PiVIT (MICCAI'23)   & 64.1 & 0.46 & $<$0.03\%
             & 71.2 & 0.18 & $<$0.01\%
             & 79.7 & 0.48 & $<$0.02\% & 664.8K \\
& RDP (TMI'24)           & 65.5 & 0.37 & \textbf{$<$0.003\%}
             & 73.1 & 0.26 & \textbf{$<$0.0002\%}
             & 80.4 & 0.11 & \textbf{$<$0.003\%} & 8922.4K \\
& IIRP (CVPR'24)      & 65.8 & 0.12 & $<$0.006\%
             & 72.2 & 0.10 & $<$0.02\%
             & 80.4 & 0.09 & $<$0.02\% & 410.1K \\
& SACB (CVPR'25)      & 65.1 & 2.08 & $<$0.06\%
             & 73.1 & 1.45 & $<$0.02\%
             & 80.3 & 1.02 & $<$0.03\% & 1105.6K \\ 
& NAP (BSPC'25)      & 66.8 & 0.30 & $<$0.2\%
             &\textbf{73.7}  & 0.28 & $<$0.03\%
             & 80.4 & 0.22 & $<$0.07\% & 501.9K \\
            \cmidrule(lr){2-12}
& \textbf{iPEAR (ours)} & \cellcolor{lightpurple}\textbf{67.9} & \cellcolor{lightpurple}0.19 & \cellcolor{lightpurple}$<$0.009\%
             & \cellcolor{lightpurple}\textbf{73.7} & \cellcolor{lightpurple}0.15 & \cellcolor{lightpurple}$<$0.003\%
             & \cellcolor{lightpurple}\textbf{80.6} & \cellcolor{lightpurple}0.12 & \cellcolor{lightpurple}$<$0.008\% & \cellcolor{lightpurple}508.7K \\ 
\cmidrule[\heavyrulewidth]{2-12}
\end{tabular}}
\label{tab:1}
\vspace{-0.4cm}
\end{table*}
\subsection{Loss Functions}\label{subsec:D}
Being an unsupervised registration network, similar to the prior SOTA model \cite{b24}, iPEAR employs NCC to evaluate the loss between the fixed image $I_f$ and the warped moving image $I_{w}^m$, which is defined as follows: 
\begin{equation}\resizebox{\linewidth}{!}{$
  \mathcal{L}_{ncc}(I_{f},I_m^w)= 
  -\sum\limits_{p\in\Omega}  \frac{\sum\limits_{p_i}(I_f(p_i)-\overline{I_f}(p))\cdot (I_m^w(p_i)-\overline{I_m^w}(p))}{\sqrt{\sum\limits_{p_i}(I_f(p_i)-\overline{I_f}(p))^2 \cdot \sum\limits_{p_i}(I_m^w(p_i)-\overline{I_m^w}(p))^2}},
$}\end{equation}
where $\Omega$ denotes the entire volume domain (i.e., the set of all voxels of the fixed image), $p_i$ denotes the $i$th voxel within $p$'s local neighborhood (i.e., the cubic patch of size $n^3$ centered at $p$), $\overline{I_f}(p)$ and $\overline{I_m^w}(p)$ denote the averaged voxel value of $I_f(p)$ and $I_m^w(p)$, respectively, computed over the corresponding neighborhoods.


Moreover, to enforce the smoothness of the deformation field, we incorporate the $L_2$-norm $|| \cdot ||$ on the gradient of $\varphi$:
\begin{equation}\mathcal{L}_{smooth}(\varphi)=\sum\nolimits_{p\in\Omega}\lVert\nabla\varphi(p)\rVert^{2}.\end{equation}
Therefore, the total loss in iPEAR is defined as follows:
\begin{equation}\mathcal{L}_{total}(I_f,I_m)=\mathcal{L}_{ncc}(I_f,I_m^w)+\lambda\mathcal{L}_{smooth}(\varphi),\end{equation}
where $\lambda$ denotes the hyperparameter used to leverage relative importance in the overall loss function.


\section{Experimental Results}

To assess the performance of iPEAR, we conduct extensive experiments on three public MRI datasets, namely Mindboggle \cite{b34}, LPBA \cite{b33}, and IXI \cite{b35}, and one public abdomen CT dataset FLARE \cite{b49}. While the SOTA models  \cite{b24},  \cite{b47}, and \cite{b48}  were evaluated using two or three datasets, ours is more extensively evaluated. Appendix~\ref{appendix1} describes these datasets, evaluation metrics, and experimental setups. 

\subsection{Comparison against Baseline Models}
\label{sec4.1}
\begin{figure*}[!t]
    \centering
\includegraphics[width=0.7\linewidth]{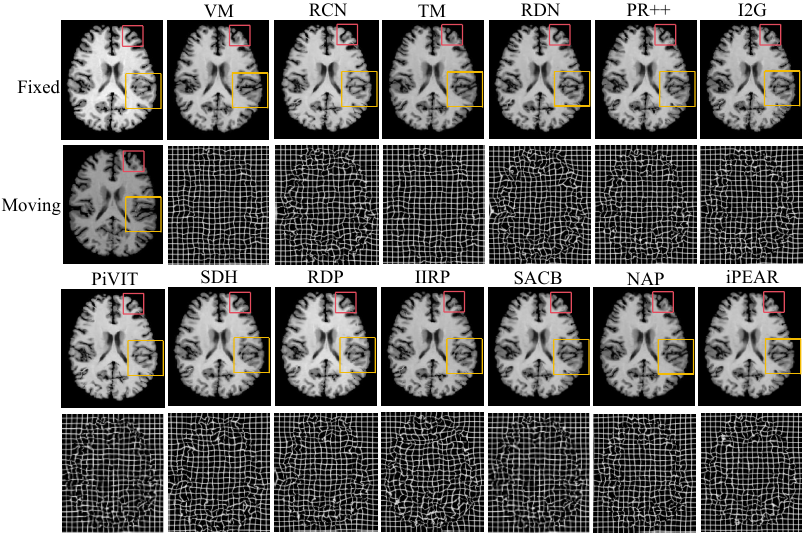}
    \caption{Visualization of registration results and corresponding deformation fields across different models on the Mindboggle dataset.}
    \label{fig:5}
 \vspace{-0.5cm}
\end{figure*}

\begin{table}[!t]
\caption{Performance comparison on the FLARE dataset}
\centering
\vspace{-0.2cm}
\setlength{\tabcolsep}{0.5mm}
\small
    \begin{tabular}{c cccc}
\specialrule{1pt}{1pt}{1pt} 
\textbf{Model} & Dice(\%)↑ & Time(s)↓ & $|{J}_{\varphi}|_{\leq0}$↓  \\
\specialrule{0.4pt}{1pt}{1pt}  
PR++ (MIA'22)    & 60.2 & 0.07 & $<$1.4\%  \\
RDP (TMI'24)    & 73.0 & 0.08 & \textbf{$<$0.03\%} \\
IIRP (CVPR'24)    & 73.8 & \textbf{0.03} & $<$2.0\% \\ 
SACB (CVPR'25)    & 74.3 & 0.13 & $<$0.7\% \\   
NAP (BSPC'25)    & 74.9 &  0.16 & $<$2.0\% \\   
\midrule
\textbf{iPEAR (ours)} & \cellcolor{lightpurple}\textbf{76.9} & \cellcolor{lightpurple}0.16 & \cellcolor{lightpurple}$<$2.0\% \\
\specialrule{1pt}{1pt}{1pt} 
\end{tabular}
\label{tab:flare} 
\vspace{-0.2cm}
\end{table}
To benchmark the performance of iPEAR, we choose two U-Net models, namely VM \cite{b11} and TM \cite{b14}, two cascaded models, namely RCN \cite{b20} and SDH \cite{b30}, and eight multi-scale pyramid models, namely RDN \cite{b17}, PR++ \cite{b31}, I2G \cite{b23}, PIVIT \cite{b50}, RDP \cite{b26}, IIRP \cite{b24}, SACB \cite{b47}, and NAP \cite{b48} as baseline models. For these baselines, we evaluate them using the same data preprocessing pipeline as iPEAR to ensure a fair comparison. To account for randomness, we evaluate each model three times with different random seeds on each dataset and report the averaged results. Unfortunately, we are not able to include certain recent baseline models (ReCorr \cite{ReCorr} and RDMR \cite{rdmr})  due to the incompleteness of the released source code. 
Table~\ref{tab:1} presents the performance of different registration models on the Mindboggle, LPBA, and IXI datasets, measured using the Dice \cite{b37}, Jacobian determinant ($|{J}_{\varphi}|_{\leq0}$), and inference time (s) metrics. Appendix~\ref{appendix5} presents HD95 \cite{b38}, ASSD \cite{b39}, and MSE metrics for these registration models across all three datasets. Table~\ref{tab:flare} presents the performance of different registration models (competing pyramid models in Table~\ref{tab:1}) on the FLARE dataset. 
The results shown in both tables demonstrate that iPEAR outperforms all the other registration models on the Dice metric across all four datasets, with particularly strong gains on Mindboggle, a well-known challenging benchmark with large spatial displacements. These gains are largely attributed to FARM, which mitigates the limitation of conventional residual convolutional decoders that are widely adopted by existing models (e.g., SACB). Such decoders treat feature channels uniformly and may miss subtle but critical deformation cues. 
Equipped with FARM, iPEAR produces more accurate and robust deformation estimates by emphasizing informative features and suppressing irrelevant ones. Furthermore, FARM is lightweight, enabling iPEAR to achieve superior performance with only a modest increase in parameter size (approximately 1.2×) over IIRP. Consequently, the inference time does not increase substantially. Unlike SACB, which estimates the deformation field only once per decoding layer, iPEAR employs TCI to adaptively determine the number of iterations at each layer, thereby yielding superior performance, achieving Dice improvements of 2.8\%, 0.6\%, 0.3\%, and 2.6\% on the four datasets, respectively. Notably, RDP and IIRP also employ iterative strategies. However, RDP adopts a fixed number of iterations and IIRP relies on a single-stage stopping criterion evaluation, both of which lead to inferior performance. In contrast, our iPEAR employs TCI with a dual-stage strategy at each decoding layer, thereby achieving elevated performance while incurring only a marginal increase in inference time over IIRP.  

Figure~\ref{fig:5} and Appendix~\ref{appendix-vis} visualize the registration results on the Mindboggle and LPBA datasets, respectively. On Mindboggle, performance differences among models are more noticeable in the two structural regions of the brain images, highlighted using the red and yellow boxes. These visualizations further demonstrate that the moving images, after being warped by the deformation fields produced by iPEAR, align more closely with fixed images. Additionally, we perform quantitative analyses of specific regions in the left and right hemispheres to further illustrate variations in anatomical structure registration performance across different models. The results, as presented in Appendix~\ref{appendix-quan}, further demonstrate the superior performance of iPEAR. 



\subsection{Generalizability of FARM and TCI}
\begin{table}[!t]
\small
\setlength{\tabcolsep}{1mm}
\centering
\caption{The generalizability of FARM on the IXI dataset}\vspace{-0.2cm}
\begin{tabular}{l cc }
\toprule
\textbf{Model} &  Dice(\%) ↑ & MSE($10^{-3}$) ↓  \\
\midrule
I2G (\textbf{+ FARM})         & 79.3 (\colorbox{lightpurple}{\textbf{79.9}}) & 0.40 (\colorbox{lightpurple}{\textbf{0.37}}) \\
\midrule
PR++  (\textbf{+ FARM})         & 79.3 (\colorbox{lightpurple}{\textbf{79.4}}) & 0.36 (\colorbox{lightpurple}{\textbf{0.32}})\\
\midrule
IIRP   (\textbf{+ FARM})       & 80.4 (\colorbox{lightpurple}{\textbf{80.5}}) & 0.24  (\colorbox{lightpurple}{\textbf{0.20}}) \\
\bottomrule
\end{tabular}
\vspace{-0.3cm}
\label{tab:dice_mse_split}
\end{table}

\begin{table}[!t]
\small
\caption{The generalizability of TCI}
\vspace{-0.2cm}
\centering
\setlength{\tabcolsep}{1mm}
\resizebox{1\linewidth}{!}{
\begin{tabular}{c ccc ccc}  
\toprule
 & \multicolumn{3}{c}{\textbf{Mindboggle}} & \multicolumn{3}{c}{\textbf{LPBA}}  \\ 
\cmidrule(lr){2-4} \cmidrule(lr){5-7}  
 & Dice(\%) & MSE↓ & Time(s)↓ & Dice(\%) & MSE↓ & Time(s)↓    \\  
\midrule
IIRP  & 65.8 & 4.16 & \textbf{0.12} & 72.2 & 0.62 & \textbf{0.10} \\  
IIRP+TCI & \cellcolor{lightpurple}\textbf{66.2} & \cellcolor{lightpurple}\textbf{3.56} & \cellcolor{lightpurple}0.14 & \cellcolor{lightpurple}\textbf{73.0} & \cellcolor{lightpurple}\textbf{0.61} & \cellcolor{lightpurple}0.14 \\  
\bottomrule
\end{tabular}}
\label{tab:6}
\vspace{-0.2cm}
\end{table}

To demonstrate the generalizability of FARM and TCI, we integrate each into existing registration models and report the results in Tables~\ref{tab:dice_mse_split} and \ref{tab:6}, respectively.  Table~\ref{tab:dice_mse_split} shows that replacing the decoder with FARM yields consistent performance gains across all three pyramid models. This is because 
FARM adaptively emphasizes anatomically relevant features while suppressing less informative ones, thereby mitigating error accumulation during coarse-to-fine decoding. Furthermore, Table~\ref{tab:6} demonstrates that substituting IIRP’s original single-stage iterative strategy with TCI yields performance improvement, underscoring the advantage of TCI in determining the number of optimization iterations dynamically. To provide in-depth details of this gain, we present the number of iterations determined by TCI and by the single-stage strategy across the four decoding levels of IIRP in Appendix~\ref{appendix-tci}. Overall, these results indicate that TCI can alleviate premature termination.

\subsection{Ablation Studies}
We further conduct ablation studies to assess the effectiveness of all components in FARM. Table~\ref{tab:2} reports ablations by enabling/disabling SEB, SAB, and RP in different combinations. Because spatial attention relies on channel-wise aggregation and should be applied on well-calibrated channel representations, we do not evaluate the variant swapping the order of SEB and SAB. It is not surprising that the model performs worst when SEB, SAB, and RP are all removed (relying solely on a single convolutional layer ${Cl}_3$).  Moreover, removing any individual module or pathway also degrades performance comparing to the intact FARM. These results suggest that SEB, SAB, and RP act synergistically, enhancing the capture of subtle anatomical structures while suppressing irrelevant features, thereby improving registration accuracy both individually and holistically.
\begin{table}[!t]
\caption{Ablation study on different FARM modules}
\centering
 \small 
 \vspace{-0.2cm}
\setlength{\tabcolsep}{1mm}
\resizebox{1\linewidth}{!}{
\begin{tabular}{ccc cc cc cc}
\toprule
\multirow{2}{*}{\textbf{SEB}} & \multirow{2}{*}{\textbf{SAB}} & \multirow{2}{*}{\textbf{RP}} & \multicolumn{2}{c}{\textbf{Mindboggle}} & \multicolumn{2}{c}{\textbf{LPBA}} & \multicolumn{2}{c}{\textbf{IXI}} \\
\cmidrule(lr){4-5} \cmidrule(lr){6-7} \cmidrule(l){8-9}
\multicolumn{3}{c}{} & Dice(\%)↑  & MSE↓ & Dice(\%)↑ & MSE↓ & Dice(\%)↑  &  MSE↓ \\
\midrule
✗ & ✗ & ✗ & 65.3 & 4.16 & 72.4 & 0.79 & 80.1 & 0.26 \\
✗ & ✗ & ✓ & 65.8 & 4.10 & 72.5 & 0.73 & 80.3 & 0.24 \\
✓ & ✗ & ✓ & 66.2 & 4.02 & 73.4 & 0.71 & 80.4 & 0.24 \\
✗ & ✓ & ✓ & 66.1 & 4.00 & 73.3 & 0.68 & 80.3 & 0.24 \\
✓ & ✓ & ✗ & 67.5 & 3.98 & 73.5 & 0.65 & 80.4 & 0.23 \\
✓ & ✓ & ✓ & \textbf{67.9} & \textbf{3.97} & \textbf{73.7} & \textbf{0.61} & \textbf{80.6} & \textbf{0.23} \\
\bottomrule
\end{tabular}}
\label{tab:2}
\vspace{-0.3cm}
\end{table}

In addition, we investigate the performance of four TCI configurations on the IXI dataset. Specifically, $\text{TCI-1}$ performs only convergence evaluation, $\text{TCI-2}$ performs only stability evaluation, $\text{TCI-3}$ evaluates convergence before stability, and $\text{TCI-4}$ evaluates stability before convergence (the configuration adopted in iPEAR). Table~\ref{tab:TCI} shows that $\text{TCI-4}$ performs best, showcasing the appropriateness of the proposed TCI design.

\begin{table}[!t]
\centering
\small 
\caption{Ablation study on different TCI configurations}
\vspace{-0.2cm}
\setlength{\tabcolsep}{1mm}
\begin{tabular}{ccccc}
\toprule
 & $\text{TCI-1}$ & $\text{TCI-2}$   & $\text{TCI-3}$  & $\text{TCI-4}$\\
\midrule
Dice(\%)↑ & 80.4  &  80.4 & 80.5& \textbf{80.6} \\
MSE($10^{-3}$)↓ & 0.232  &  0.248 & 0.231 &\textbf{0.231}\\
\bottomrule
\end{tabular}
\label{tab:TCI}
\vspace{-0.2cm}
\end{table}
Finally, Appendices~\ref{appendixtic}$\sim$\ref{appendix4} assess the effectiveness of TCI across different decoding layers and the influence of varying values of thresholds $\delta_s$,  $\delta_c$ and $k_{max}$, the choice of similarity metric, and the window size $t$, respectively.



\section{Conclusion}
We propose the deformable image registration network iPEAR, which employs FARM to extract anatomical structure details and learn to suppress irrelevant features. Additionally, we introduce the dual-stage TCI strategy to determine the number of optimization iterations. The results on four datasets demonstrate the effectiveness of iPEAR. Generalizability and ablation studies validate the effectiveness of FARM and TCI. Currently, we combine channel attention and spatial attention to suppress irrelevant features. Going forward, we believe further improvement is achievable by extending this to a multi-head attention mechanism to enhance the model's capability in feature representation.



\bibliography{iPEAR}
\bibliographystyle{icml2026}

\newpage
\appendix
\onecolumn

\setcounter{figure}{0} 
\renewcommand{\thefigure}{\thesection} 
\setcounter{table}{0}
\renewcommand{\thetable}{\thesection}

\section{Datasets and Implementation Details}
\label{appendix1}
\textbf{Datasets:} Consistent with prior studies \cite{b48,b26}, iPEAR focuses on single-modality registration. Accordingly, to comprehensively assess the performance of the proposed iPEAR, we compare it against different registration models on three public MRI datasets, namely Mindboggle \cite{b34}, LPBA \cite{b33},  and IXI \cite{b35}, and one public abdomen CT dataset FLARE \cite{b49}. We preprocess these four datasets as follows.  For the Mindboggle dataset, we choose the NKI-RS-22 and NKI-TRT-20 sub-datasets from Mindboggle101 as the training set and OASIS-TRT-20 as the test set, following the prior study \cite{b24}. We crop each volume to  $160\times192\times160$ with an isotropic voxel size of 1 mm, while all images are pre-aligned with the MNI 152 template space. The LPBA dataset contains 40 T1-weighted MRI volumes, each annotated with 54 manually labeled Regions of Interest (ROIs). All volumes have been rigorously pre-aligned to the MNI 305. Specifically, each volume is cropped to $160\times192\times 160$ voxels with an isotropic voxel size of 1 mm. We use 31 volumes for training and 9 for testing, following the prior study \cite{b26}. The IXI dataset contains the original brain MRI images provided by \cite{b36} and their corresponding 3D volume data after affine registration. All volumes are resampled to a voxel size of $80\times96\times112$ (isotropic voxel size of 2 mm), and for training, we use 407 3D volumes (total $407\times406$ paired samples), and for testing, 58 volumes (total $58\times57$ paired samples), following the prior study \cite{b26}. The FLARE dataset consists of 361 scans annotated with four organ labels, namely liver, kidney, spleen, and pancreas. Each volume is cropped to $128\times128\times96$ voxels with an isotropic voxel size of 2.5 mm. We allocate 341 volumes for training and 20 for testing following the \cite{b24}. 

\noindent\textbf{Metrics:} To comprehensively assess the performance of various registration models, we adopt multiple quantitative metrics. The Dice score \cite{b37} measures region‑wise overlap, while the 95\% maximum Hausdorff Distance (HD95) \cite{b38} and the Average Symmetric Surface Distance (ASSD) \cite{b39} quantify the similarity of the regional profiles. The quality of the estimated deformation field $\varphi$ is assessed by the percentage of voxels with non-positive Jacobian determinant $|{J}_{\varphi}|_{\leq0}$↓ (i.e., folded voxels). In addition, we report Mean Squared Error (MSE), Inference Time (s), and model size (the number of parameters) for a comprehensive performance analysis. 

\noindent\textbf{Implementation Details:} Our Model is implemented using Pytorch and trained on an NVIDIA GeForce RTX 4090 GPU. Following the prior study \cite{b24}, we normalize all images to the range $[0, 1]$, optimize using the ADAM optimizer \cite{b40} at a learning rate of $1\times 10^{-4}$, and set the size of volumetric patch $n$, hyperparameter $\lambda$, and batch size to 27, 1, and 1, respectively. In FARM, we adopt a reduction ratio $r=16$. In TCI, we set stability threshold $\delta_s$, convergence threshold  $\delta_c$, the window size $t$, and the maximum number of iterations $k_{max}$ to 0.005, 0.005, 3, and 10, respectively. 
\section{Visualization of Feature Attention Patterns for AP and RP}
\label{appendix_grad}
\begin{figure*}[!t]
    \centering
    \includegraphics[width=0.9\linewidth]{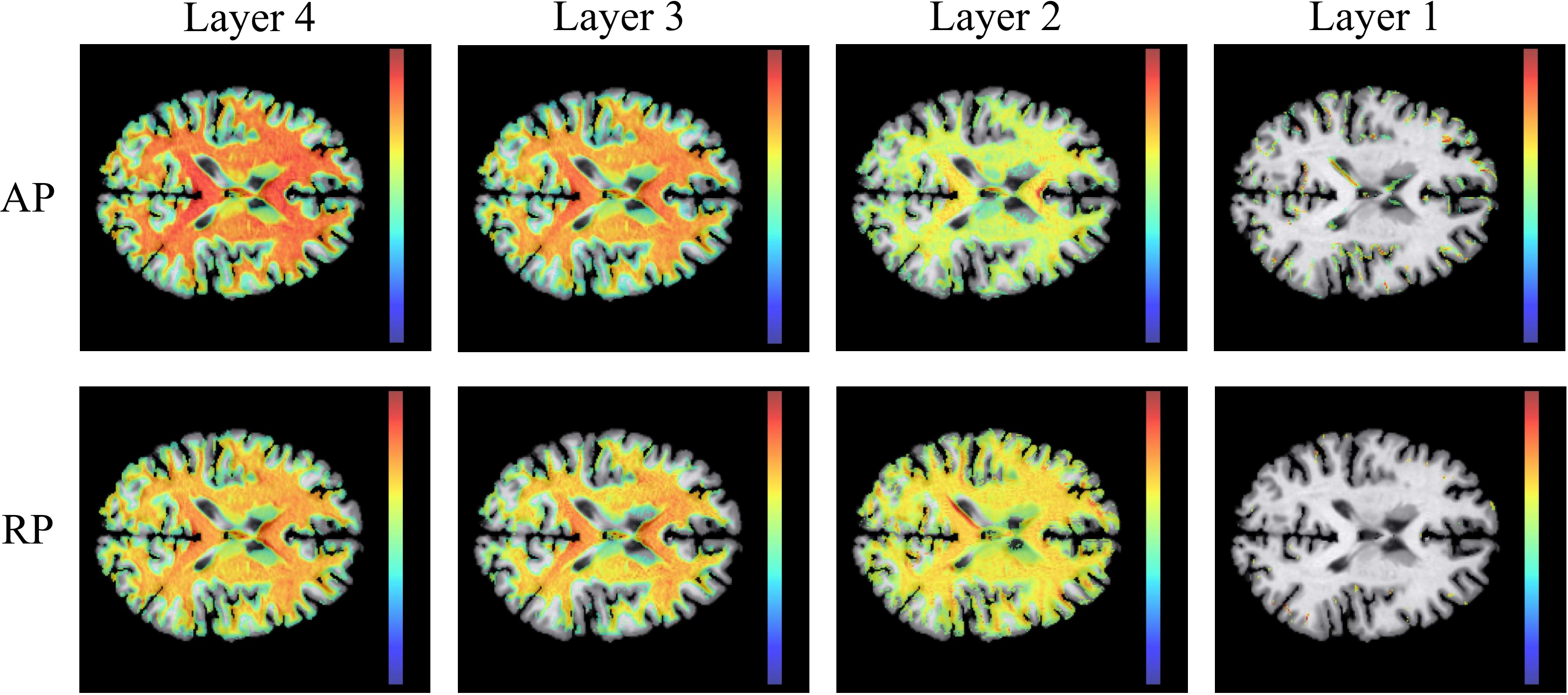}
    \caption{Visualization of Feature Attention Patterns for AP and RP.}
    \label{fig:grad-cam}
\end{figure*} 
As shown in Figure~\ref{fig:grad-cam}, we illustrate the distinct feature attention patterns of AP and RP at each decoding Layer. The AP heatmaps concentrate more strongly on core registration regions (e.g., central gray matter), exhibiting a markedly higher response density, whereas the RP responses are more uniformly distributed. In addition, at deep decoder layers (e.g., $4$th Layer, corresponding to the coarse registration stage), large global structural discrepancies between the warped and fixed images encourage the network to exploit multi-regional features, resulting in widespread high-response regions in the heatmaps. As decoding proceeds from deep to shallow layers, registration errors decrease and anatomical structures become increasingly aligned, so the model primarily focuses on local regions for fine-grained adjustments. These visualizations demonstrate the rationality of our dual-pathway design.

\section{Performance Comparison on HD95, ASSD, and MSE Metrics}
\label{appendix5}
\begin{table*}[!t]
\centering
\caption{Performance comparison of different registration models on the Mindboggle, LPBA, and IXI datasets evaluated by HD95, ASSD, and MSE metrics}
\renewcommand{\arraystretch}{1.2}
\setlength{\tabcolsep}{3.5pt}
\small
\resizebox{0.9\linewidth}{!}{
    \begin{tabular}{c c ccc ccc ccc}
\cmidrule[\heavyrulewidth]{2-11}
& \multirow{2}{*}{\textbf{Model}} & \multicolumn{3}{c}{\textbf{Mindboggle}} & \multicolumn{3}{c}{\textbf{LPBA}} & \multicolumn{3}{c}{\textbf{IXI}} \\
\cmidrule(lr){3-5}
\cmidrule(lr){6-8}
\cmidrule(lr){9-11}
& & HD95↓ & ASSD↓ & MSE($10^{-3}$)↓
  & HD95↓ & ASSD↓ & MSE($10^{-3}$)↓ 
  & HD95↓ & ASSD↓ & MSE($10^{-3}$)↓  \\ \cmidrule(lr){2-11}
\multirow{2}{*}{\rotatebox[origin=c]{90}{U-Net}} 
& VM (CVPR'18)    & 5.683 & 1.726 & 7.449 
             & 6.557 & 2.089 & 2.293 
             & 1.460 & 0.584 & 0.727   \\
& TM (MIA'22)    & 5.399 & 1.561 & 5.500 
             & 6.439 & 1.962 & 1.179 
             & 1.432 & 0.557 & 0.432   \\ \cmidrule(lr){2-11}
\multirow{2}{*}{\rotatebox[origin=c]{90}{Cascade}} 
& RCN (ICCV'19)         & 5.090 & 1.439 & 4.607 
             & 5.817 & 1.777 & 2.005 
             & 1.397 & 0.563 & 0.518   \\         
& SDH-Net (TMI'23)      & 5.047 & 1.413 & 4.616 
             & 5.750 & 1.682 & 0.963 
             & 1.362 & 0.510 & 0.392   \\ \cmidrule(lr){2-11}
\multirow{8}{*}{\rotatebox[origin=c]{90}{Pyramid}} 
& RDN (JBHI'22)           & 5.041 & 1.468 & 5.884 
             & 5.817 & 1.747 & 1.365 
             & 1.270 & 0.512 & 0.570   \\
& PR++ (MIA'22)  & 5.312 & 1.510 & 4.871 
             & 6.060 & 1.778 & 0.852 
             & 1.290 & 0.508 & 0.360   \\
& I2G (MMMI'22)   & 5.129 & 1.474 & 4.347 
             & 5.646 & 1.665 & 1.018 
             & 1.301 & 0.509 & 0.408   \\
& PiVIT (MICCAI'23)   & 5.012 & 1.443 & 4.297 
             & 5.630 & 1.647 & 1.001 
             & 1.268 & 0.487 & 0.336   \\
& RDP (TMI'24)           & 4.992 & 1.392 & 4.191 
             & 5.618 & 1.621 & 0.633 
             & 1.229 & 0.472 & 0.234   \\
& IIRP (CVPR'24)      & 4.947 & 1.393 & 4.163 
             & 5.737 & 1.605 & 0.623 
             & \textbf{1.225} & 0.476 & 0.236   \\
& SACB (CVPR'25)      & 4.884 & 1.386 & 4.852 
             & 5.862 & \textbf{1.326} & 0.837 
             & 1.297 & 0.504 & 0.351   \\ 
& NAP (BSPC'25)      & \textbf{4.832} & 1.373 & 4.031 
             & 5.570 & 1.384 & 0.707 & 
             1.228 & 0.470 & 0.233    \\\cmidrule(lr){2-11}
& \textbf{iPEAR (ours)} & \cellcolor{lightpurple}4.912 & \cellcolor{lightpurple}\textbf{1.366} & \cellcolor{lightpurple}\textbf{3.972} 
             & \cellcolor{lightpurple}\textbf{5.569} & \cellcolor{lightpurple}1.652 & \cellcolor{lightpurple}\textbf{0.609} 
             & \cellcolor{lightpurple}1.231 & \cellcolor{lightpurple}\textbf{0.460} & \cellcolor{lightpurple}\textbf{0.231}   \\
\cmidrule[\heavyrulewidth]{2-11}
\end{tabular}}
\label{tab:other metrics}
\end{table*}

In Table~\ref{tab:other metrics}, we present the performance of different registration models on the Mindboggle, LPBA, and IXI datasets, evaluated by the HD95, ASSD, and MSE metrics. The experimental results demonstrate that iPEAR outperforms the other registration models in most scenarios and achieves SOTA results on at least two metrics for each dataset.

\section{Visualization of Registration Results on the LPBA Dataset}
\label{appendix-vis}
\begin{figure*}[!t]
    \centering
    \includegraphics[width=0.9\linewidth]{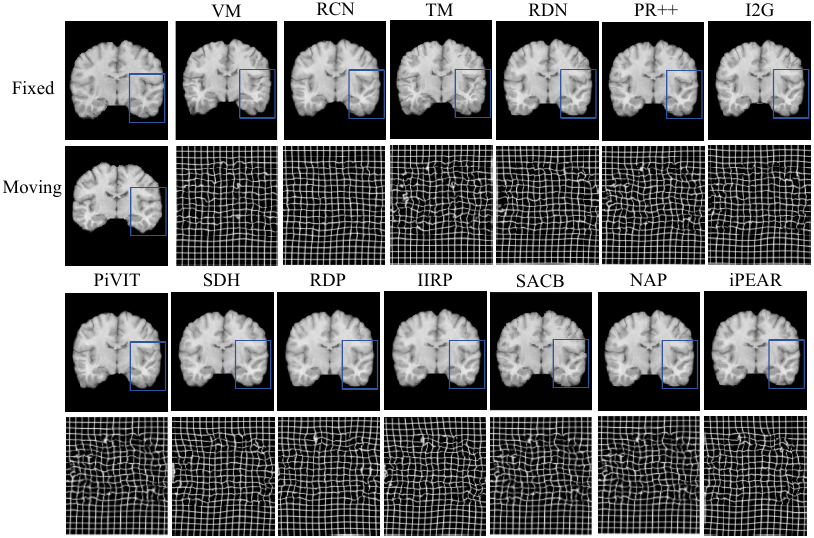}
    \caption{Visualization of registration results and corresponding deformation fields across different models on the LPBA dataset.}
    \label{fig:LPBA}
\end{figure*}
In Figure~\ref{fig:LPBA}, we present the registration results and corresponding deformation fields across different models on the LPBA dataset. On LPBA, performance differences among models are more noticeable in the structural region of the brain images, highlighted using the blue box. These visualizations further show that, after warping with the deformation fields estimated by iPEAR, the moving images exhibit closer alignment with the fixed images.

\setcounter{figure}{0} 
\renewcommand{\thefigure}{\thesection.\arabic{figure}}

\section{Quantitative Analysis on Different Anatomical Structures}
\label{appendix-quan}
\begin{figure}[!t]
    \centering
\includegraphics[width=0.9\linewidth]{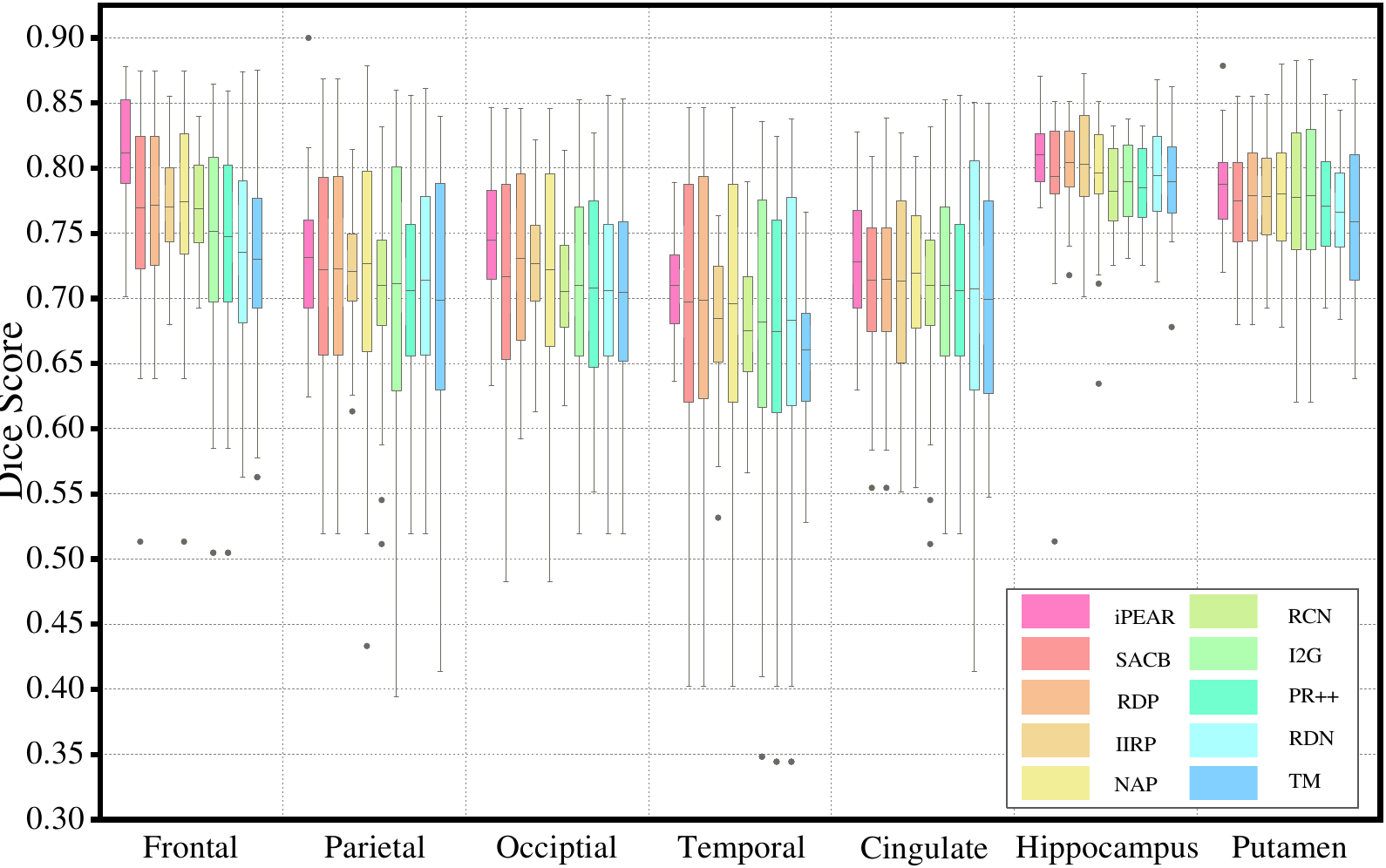}
    \caption{Performance comparison of different models in seven regions of the LPBA dataset.}
    \label{fig:6}
\end{figure}
\begin{figure}[!t]
    \centering
\includegraphics[width=0.9\linewidth]{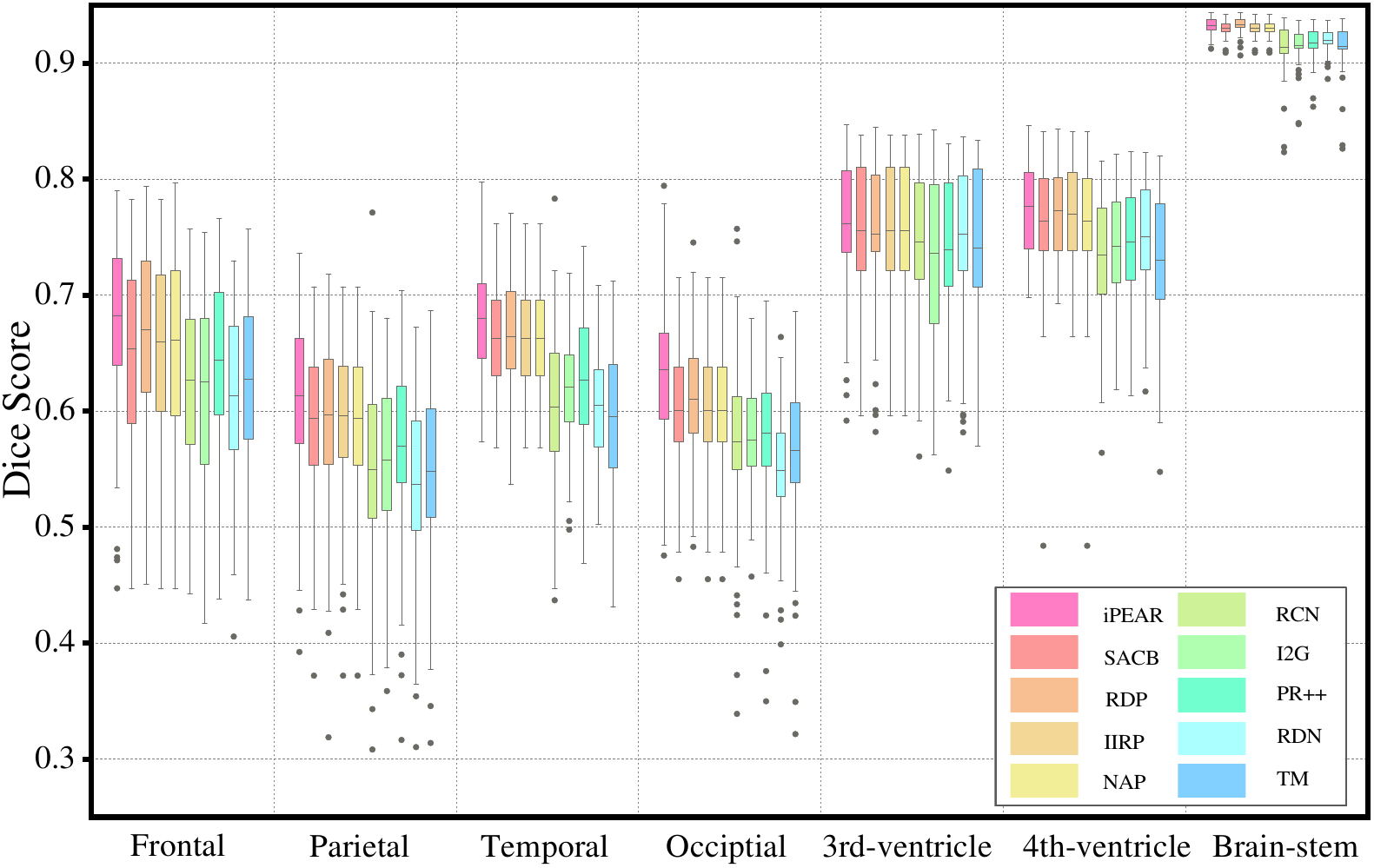}
    \caption{Performance comparison of different models in seven regions of the Mindboggle dataset.}
    \label{fig:7}
\end{figure}
\label{appendix2}
\begin{figure*}[!t]
    \centering
\includegraphics[width=0.9\linewidth]{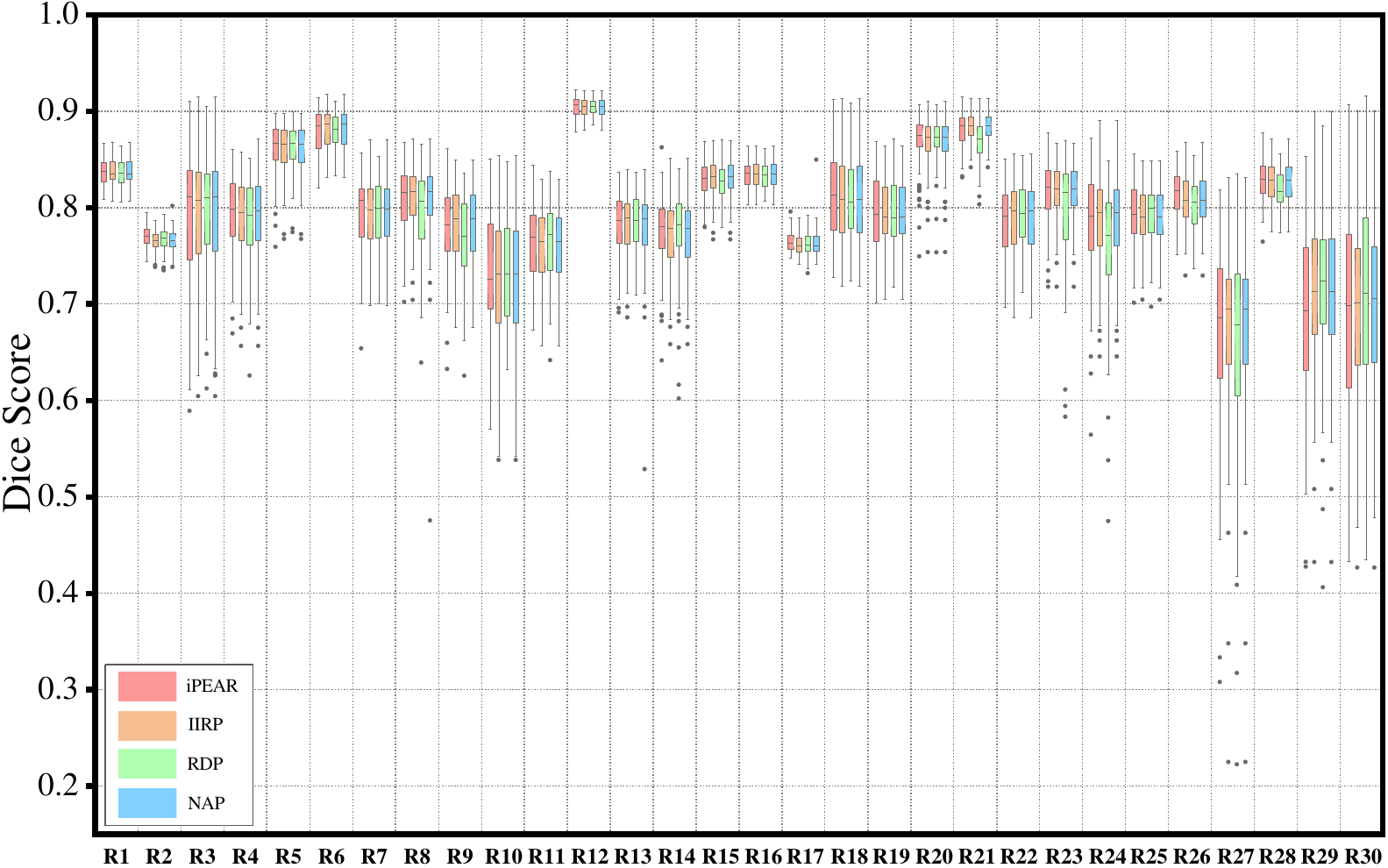}
    \caption{Performance comparison of different models in thirty regions of the IXI dataset.}
    \label{fig:8}
\end{figure*}
\renewcommand{\thefigure}{\thesection.\arabic{figure}}
Following the prior study \cite{b26}, we evaluate iPEAR’s registration performance on selected regions in both hemispheres to assess its accuracy across diverse anatomical structures.
Specifically, Figure \ref{fig:6} presents the Dice scores for seven brain regions in the LPBA dataset, including the frontal, parietal, temporal, occipital, cingulate, hippocampus, and putamen. Figure \ref{fig:7} presents the Dice scores for seven brain regions in the Mindboggle dataset, including the frontal, parietal, temporal, occipital, third ventricle, fourth ventricle, and brainstem. Figure \ref{fig:8} presents the Dice score across 30 regions of the IXI dataset. All these experimental results demonstrate that iPEAR outperforms existing registration models. 

\section{Performance Comparison  of Different Iteration Strategies}
\label{appendix-tci}
\begin{table}[!t]
\small
\caption{Performance comparison  of different iteration strategies on the Mindboggle and FLARE datasets}
\centering
\setlength{\tabcolsep}{1mm}
\begin{tabular}{c ccc ccc}  
\toprule
 & \multicolumn{3}{c}{\textbf{Mindboggle}} & \multicolumn{3}{c}{\textbf{FLARE}}  \\ 
\cmidrule(lr){2-4} \cmidrule(lr){5-7}  
 & Dice(\%)↑ & MSE($10^{-3}$)↓ & Iter & Dice(\%)↑ & MSE($10^{-3}$)↓ & Iter  \\  
\midrule
IIRP  & 65.8 & 4.16 & [1.2, 3.5, 2.0, 2.0] & 73.8 & 3.59 & [1.3, 3.4, 3.0, 2.6] \\  
IIRP (Iter = 2)  & 65.8 & 4.15 & [2.0, 2.0, 2.0, 2.0] & 73.6 & 3.66 & [2.0, 2.0, 2.0, 2.0] \\  
IIRP (Iter = 3)  & 66.0 & 4.03 & [3.0, 3.0, 3.0, 3.0] & 74.1 & 3.50 & [3.0, 3.0, 3.0, 3.0] \\  
IIRP (Iter = 4)  & 65.5 & 4.49 & [4.0, 4.0, 4.0, 4.0] & 73.7 & 3.61 & [4.0, 4.0, 4.0, 4.0] \\  
IIRP + TCI & \cellcolor{lightpurple}\textbf{66.2} & \cellcolor{lightpurple}\textbf{3.56} & \cellcolor{lightpurple}[3.9, 3.5, 3.2, 3.0] & \cellcolor{lightpurple}\textbf{74.6} & \cellcolor{lightpurple}\textbf{3.40} & \cellcolor{lightpurple}[3.0, 3.4, 3.6, 3.0] \\  
\bottomrule
\end{tabular}
\label{tab:iter_of_TCI}
\end{table}
In this section, we examine the effectiveness of TCI in mitigating premature termination, and report the number of iterations determined by TCI and by the single-stage stopping strategy, respectively. As shown in Table~\ref{tab:iter_of_TCI}, on the Mindboggle and FLARE datasets, TCI judiciously increases the average number of iterations across all decoding levels, with the largest increase at $4$th Layer. The average number of iterations rises from 1.2 to 3.9 and from 1.3 to 3.0 on the two datasets, respectively. This deeper iterative refinement reduces the risk of early-stage under-optimization and helps limit the accumulation of anatomical misalignments.
In addition, we compare TCI with a fixed-iteration setting, and the results show that TCI yields better registration performance than using a fixed number of iterations.

\section{Effectiveness of TCI at Each Decoding Layer}
\label{appendixtic}
\begin{table}[!t]
\centering
\caption{Ablation study on the effectiveness of TCI on various decoding layers}
\small
\setlength{\tabcolsep}{1mm}
\begin{tabular}{cccc|ccc}
\toprule
\textbf{Layer 4} & \textbf{Layer 3} & \textbf{Layer 2} & \textbf{Layer 1} & \textbf{Mindboggle} & \textbf{LPBA} & \textbf{IXI} \\
\midrule
 &       &       &       & 64.9 & 71.7 & 79.8 \\
         ✓ & &  &       & 65.9& 72.9 & 79.7 \\
          &   ✓    &  &  & 65.8 & 72.7 & 79.8 \\
 &       &  ✓     &  & 65.8 & 73.0 & 79.9 \\
          &  &  & ✓ & 66.0 & 73.1 & 79.9 \\
✓ & ✓      &  &  & 66.2 & 73.4 & 80.0 \\
✓ & ✓ & ✓      & & 67.3 & 73.6 & 80.2 \\
✓ & ✓ & ✓ & ✓ & \textbf{67.9} & \textbf{73.7} & \textbf{80.6} \\
\bottomrule
\end{tabular}

\label{tab:3}
\end{table}

In this section, we conduct ablation studies to assess the effectiveness of TCI at each decoding layers. Omitting TCI at a given layer fixes the number of iterations to one, thereby preventing iterative refinement of the deformation field. As shown in Table \ref{tab:3}, progressively enabling TCI across additional layers steadily improves iPEAR’s performance. Notably, applying TCI at higher-resolution layers (e.g., Layer 1) yields larger performance gains than at lower-resolution layers (e.g., Layer 4). A plausible reason for this is that higher-resolution feature maps more closely match the original image dimensions,  preserving fine-grained structural details. Consequently, determining the number of iterations for higher-resolution layers yields greater performance improvements.

\section{Impact of Varying Values of Thresholds}
\label{appendixthreshold}

\begin{table}[!t]
\centering
\caption{Ablation study on thresholds $\delta_s$ and $\delta_c$ values}
\begin{tabular}{c cc cc cc}
\toprule
 & \multicolumn{2}{c}{\textbf{Mindboggle}} & \multicolumn{2}{c}{\textbf{LPBA}} & \multicolumn{2}{c}{\textbf{IXI}} \\
\cmidrule(lr){2-3} \cmidrule(lr){4-5} \cmidrule(l){6-7}
 \diagbox[width=1.8cm,height=0.5cm]{$\delta_s$}{$\delta_c$}& $0.01$ & $0.005$ & $0.01$ & $0.005$ & $0.01$ & $0.005$ \\
\midrule
$0.01$  & 65.9 & 66.3 & 73.1& 73.2 & 80.4 & 80.4 \\
$0.005$ & 66.8 & \textbf{67.9} & 73.4 & \textbf{73.7} & 80.3 & \textbf{80.6} \\
$0.001$ & 67.0 & 67.6 & 73.3 & 73.4 & 80.3 & 80.2 \\
\bottomrule
\end{tabular}
\label{tab:4}
\end{table}

\setcounter{figure}{0}

\begin{figure}[!t]
    \centering
\includegraphics[width=0.5\linewidth]{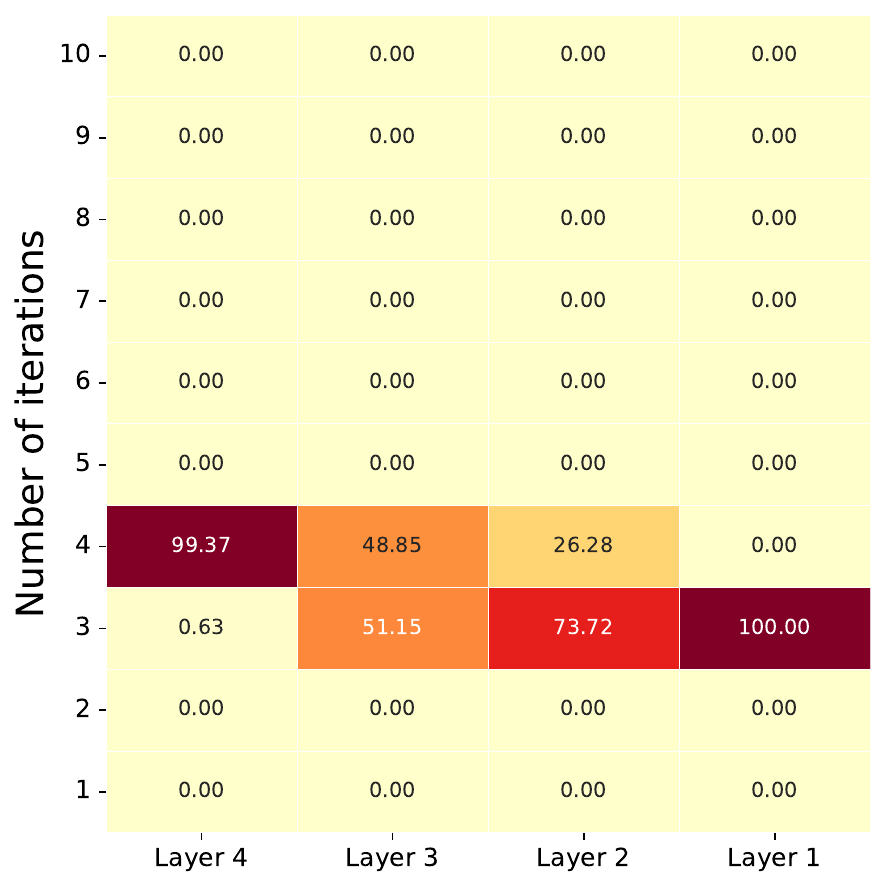}
    \caption{Statistics of the number of iterations across different decoding layers.}
    \label{fig:iteraion}
\end{figure}
In this section, we investigate the sensitivity of the predetermined thresholds $\delta_s$ and $\delta_c$. As shown in Table~\ref{tab:4}, these two thresholds exhibit low sensitivity to variation and have minimal impact on the overall performance of iPEAR.
Nonetheless, we choose the middle range values, i.e., $\delta_s=0.005$ and $\delta_c=0.005$, which yield the best performance


In addition, in Figure~\ref{fig:iteraion}, we present the number of iterations per decoding layer determined by TCI on the IXI dataset. The experimental results demonstrate that three or four iterations suffice for convergence in most cases. Hence, our predefined maximum number of iterations $k_{max} = 10$ is reasonable.

\section{Impact of Different TCI Similarity Metrics}
\label{appendix3}

In this section, we investigate how three similarity metrics, namely Normalized Cross-Correlation (NCC), Mean Absolute Error (MAE), and Mean Squared Error (MSE), affect the performance of TCI on the IXI dataset. As shown in Table~\ref{tab:10}, NCC delivers the best results. Hence, we adopt NCC as the similarity metric for all experiments in this paper.

\begin{table}[!t]
\centering
\caption{Ablation study on the effectiveness of different similarity metrics adopted in TCI}
\begin{tabular}{c cc cc cc}
\toprule
 & \multicolumn{2}{c}{\textbf{MAE}} & \multicolumn{2}{c}{\textbf{MSE}} & \multicolumn{2}{c}{\textbf{NCC}} \\
\cmidrule(lr){2-3} \cmidrule(lr){4-5} \cmidrule(l){6-7}
 \diagbox[width=1.8cm,height=0.5cm]{$\delta_s$}{$\delta_c$}& $0.01$ & $0.005$ & $0.01$ & $0.005$ & $0.01$ & $0.005$ \\
\midrule
$0.01$  & 79.6 & 79.7 & 80.0 & 79.5 & 80.4 & 80.6 \\
$0.005$ & 79.7 & 80.0 & 79.9 & 80.1 & 80.6 & \textbf{80.6} \\
$0.001$ & 80.2 & 79.1 & 80.1 & 79.9 & 80.4 & 80.2 \\
\bottomrule
\end{tabular}
\label{tab:10}
\end{table}



\section{Impact of Different Sliding Window Sizes}
\label{appendix4}
\begin{table}[!t]
 \caption{Ablation study on the different window sizes}
 \setlength{\tabcolsep}{3.5pt}
\centering
\begin{tabular}{c  ccccc ccccc}
\toprule
 & \multicolumn{5}{c}{\textbf{IXI}}& \multicolumn{5}{c}{\textbf{Mindboggle}} \\
\cmidrule(lr){2-6} \cmidrule(lr){7-11} 
 $t$ & Dice(\%)↑ & HD95↓ & ASSD↓ & MSE($10^{-3}$)↓ & Time(s)↓ &  Dice(\%)↑ & HD95↓ & ASSD↓ & MSE($10^{-3}$)↓ & Time(s)↓ \\
\midrule
3 & 80.6 & 1.23 & 0.46 & 0.23 & \textbf{0.12}  & \textbf{67.9} & \textbf{4.91} & 1.37 & 3.97 & \textbf{0.19}  \\
4 & \textbf{80.9} & \textbf{1.22} & \textbf{0.46} & \textbf{0.21} & 0.15  & 67.4 & 4.93 & 1.38 & \textbf{3.89} & 0.22\\
5 & 80.7 & 1.27 & 0.49 & 0.23 & 0.29  & 67.0 & 4.94 & 1.39 & 4.06 & 0.35 \\ 
\bottomrule
\end{tabular}
\label{tab:window_size_comparison}
\end{table}
In Table~\ref{tab:window_size_comparison}, we present the effect of window size $t$ on the performance of iPEAR. On the IXI dataset, increasing $t$ from 3 to 4 yields marginal performance gains but incurs longer inference time. Increasing $t$ further to 5 degrades performance across all metrics, suggesting that  $t=4$ provides sufficient iterative refinement without incurring excessive iterations. On the Mindboggle dataset, setting $t=3$ outperforms $t=4$ on most metrics, likely because $t=4$ induces excessive iterations in TCI’s deformation refinement. Therefore, we set $t=3$ for all experiments in this paper.



\end{document}